\documentclass[runningheads]{llncs}
\usepackage{graphicx}
\usepackage{amsmath,amssymb} 
\usepackage{color}
\usepackage[width=122mm,left=12mm,paperwidth=146mm,height=193mm,top=12mm,paperheight=217mm]{geometry}

\begin{document}
	\title{Deep High Dynamic Range Imaging with Large Foreground Motions}
	
	\titlerunning{Deep HDR with Large Foreground Motions}
	%
	\author{Shangzhe Wu\inst{1,3}\thanks{This work was partially done when Shangzhe Wu was an intern at Tencent Youtu.} \and
		Jiarui Xu\inst{1} \and
		Yu-Wing Tai\inst{2} \and
		Chi-Keung Tang\inst{1}}
	%
	\authorrunning{S. Wu, J. Xu, Y.-W. Tai and C.-K. Tang}
	%
	
	\institute{The Hong Kong University of Science and Technology \and
		Tencent Youtu \and
		University of Oxford\\
		\email{\{swuai,jxuat\}@connect.ust.hk, yuwingtai@tencent.com, cktang@cs.ust.hk}}
	\maketitle              
	\begin{abstract}
		This paper proposes the first non-flow-based deep framework for high dynamic range (HDR) imaging of dynamic scenes with \textbf{large-scale foreground motions}. In state-of-the-art deep HDR imaging, input images are first aligned using optical flows before merging, which are still error-prone due to occlusion and large motions. In stark contrast to flow-based methods, we formulate HDR imaging as an image translation problem \textbf{without optical flows}. Moreover, our simple translation network can automatically hallucinate plausible HDR details in the presence of total occlusion, saturation and under-exposure, which are otherwise almost impossible to recover by conventional optimization approaches. Our framework can also be extended for different reference images. We performed extensive qualitative and quantitative comparisons to show that our approach produces excellent results where color artifacts and geometric distortions are significantly reduced compared to existing state-of-the-art methods, and is robust across various inputs, including images without radiometric calibration. 
		
		\keywords{High Dynamic Range Imaging \and Computational Photography}
	\end{abstract}

	\begin{figure}[h]
		\begin{center}
			\footnotesize
			\includegraphics[width=\linewidth]{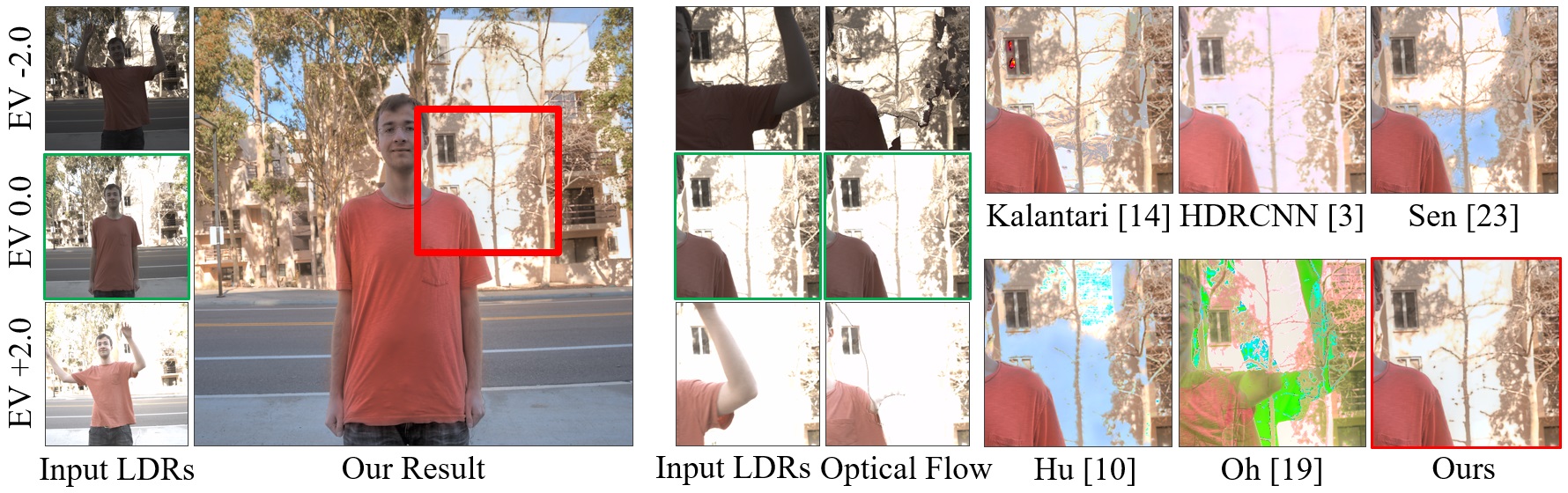}
		\end{center}
		\caption{Our goal is to produce an HDR image from a stack of LDR images that can be corrupted by large foreground motions, such as images shown on the left. Our resulted HDR image is displayed after tonemapping. On the right, the first two columns show that the optical flow alignment used by Kalantari~\cite{LearningHDR} introduces severe geometric distortions and color artifacts, which are unfortunately preserved in the final HDR results. The last three columns compare the results produced by other state-of-the-art methods and ours where no optical flow alignment is used. Our simple network produces high quality ghost-free HDR image in the presence of large-scale saturation and foreground motions.}
		\label{fig:teaser}
	\end{figure}
	
	\section{Introduction}
	Off-the-shelf digital cameras typically fail to capture the entire dynamic range of a 3D scene. In order to produce high dynamic range (HDR) images, custom captures and special devices have been proposed~\cite{Tocci11,Heide14,CSHDR_EG2016}. Unfortunately, they are usually too heavy and/or too expensive for capturing fleeting moments to cherish, which are typically photographed using cellphone cameras. The other more practical approach is to merge several low dynamic range (LDR) images captured at different exposures. If the LDR images are perfectly aligned, in other words no camera motion or object motion is observed, the merging problem is considered almost solved~\cite{Mann95,Debevec97}. However, foreground and background misalignments are unavoidable in the presence of large-scale foreground motions in addition to small camera motions. While the latter can be resolved to a large extent by homography transformation~\cite{Tomaszewska07}, foreground motions, on the other hand, will make the composition nontrivial. Many existing solutions tackling this issue are prone to introducing artifacts or ghosting in the final HDR image~\cite{Kang03,Zimmer11,LearningHDR}, or fail to incorporate misaligned HDR contents by simply rejecting the pixels in misaligned regions as outliers~\cite{Khan06,Heo2011,Oh15}, see Fig.~\ref{fig:teaser}. 
	
	Recent works have been proposed to learn this composition process using deep neural networks~\cite{LearningHDR}. In~\cite{LearningHDR}, they first used optical flow to align input LDR images, followed by feeding the aligned LDRs into a convolutional neural network (CNN) to produce the final HDR image. Optical flows are often unreliable, especially for images captured with different exposure levels, which inevitably introduce artifacts and distortions in the presence of large object motions. Although in~\cite{LearningHDR} it was claimed that the network is able to resolve these issues in the merging process, failure cases still exist as shown in Fig.~\ref{fig:teaser}, where color artifacts and geometry distortions are quite apparent in the final results. 
	
	In contrast, we regard merging multiple exposure shots into an HDR image as an image translation problem, which have been actively studied in recent years. In~\cite{pix2pix} a powerful solution was proposed to learn a mapping between images in two domains using a Generative Adversarial Network (GAN). Meanwhile, CNNs have been demonstrated to have the ability to learn misalignment~\cite{flownet15} and hallucinate missing details~\cite{hallucination}. Inspired by these works, we believe that optical flow may be an overkill for HDR imaging. In this paper, we propose a simple end-to-end network that can learn to translate multiple LDR images into a ghost-free HDR image even in the presence of large foreground motions. 
	
	In summary, our method has the following advantages. First, unlike~\cite{LearningHDR}, our network is trained end-to-end without optical flow alignment, thus intrinsically avoiding artifacts and distortions caused by erroneous flows. In stark contrast to prevailing flow-based HDR imaging approaches~\cite{LearningHDR}, this provides a novel perspective and significant insights for HDR imaging, and is much faster and more practical. Second, our network can hallucinate plausible details that are totally missing or their presence is extremely weak in all LDR inputs. This is particularly desirable when dealing with large foreground motions, because usually some contents are not captured in all LDRs due to saturation and occlusion. Finally, the same framework can be easily extended to more LDR inputs, and possibly with any specified reference image. We perform extensive qualitative and quantitative comparisons, and show that our simple network outperforms the state-of-the-art approaches in HDR synthesis, including both learning based or optimization based methods. We also show that our network is robust across various kinds of input LDRs, including images with different exposure separations and images without correct radiometric calibration. 
	
	\section{Related Work}
	Over the past decades, many research works have been dedicated to the problem of HDR imaging. As mentioned above, one practical solution is to compose an HDR image from a stack of LDR images. Early works such as~\cite{Mann95,Debevec97} produce excellent results for static scenes and static cameras.
	
	To deal with camera motions, previous works~\cite{Kang03,Tomaszewska07,Jacobs08} register the LDR images before merging them into the final HDR image. Since many image registration algorithms depend on the brightness consistence assumptions, the brightness changes are often addressed by mapping the images to another domain, such as luminance domain or gradient domain, before estimating the transformation. 
	
	Compared to camera motions, object motions are much harder to handle. A number of methods reject the moving pixels using weightings in the merging process~\cite{Khan06,Heo2011}. Another approach is to detect and resolve ghosting after the merging~\cite{Gallo09,Raman2011}. Such methods simply ignore the misaligned pixels, and fail to fully utilize available contents to generate an HDR image. 
	
	There are also more complicated methods~\cite{Kang03,Zimmer11} that rely on optical flow or its variants to address dense correspondence between image pixels. However, optical flow often results in artifacts and distortions when handling large displacements, introducing extra complication in the merging step. Among the works in this category,~\cite{LearningHDR} produces perhaps the best results, and is highly related to our work. The authors proposed a CNN that learns to merge LDR images aligned using optical flow into the final HDR image. Our method is different from theirs in that we do not use optical flow for alignment, which intrinsically avoids the artifacts and distortions that are present in their results. We provide concrete comparisons in the later sections. 
	
	Another approach to address the dense correspondence is patch-based system~\cite{Sen12,Hu13}. Although these methods produce excellent results, the running time is much longer, and often fail in the presence of large motions and large saturated regions. 
	
	A more recent work~\cite{hdrcnn} attempts to reconstruct a HDR image from one single LDR image using CNN. Although their network can hallucinate details in regions where input LDRs exhibit only very weak response, one intrinsic limitation of their approach is the total reliance on one single input LDR image, which often fails in highly contrastive scenes due to large-scale saturation. Therefore, we intend to explore better solutions to merge HDR contents from multiple LDR images, which can easily be captured in a burst, for instance, using cellphone cameras.
	
	Typically, to produce an HDR image also involves other processing, including radiometric calibration, tone-mapping and dynamic range compression. Our work is focused on the merging process. Besides, there are also more expensive solutions that use special devices to capture a higher dynamic range~\cite{Tocci11,Heide14,CSHDR_EG2016} and directly produce HDR images. For a complete review of the problem, readers may refer to~\cite{Gallo201685}. 
	
	\begin{figure}[t]
		\begin{center}
			\footnotesize
			\begin{tabular}{c c}
				\includegraphics[width=0.70\textwidth]{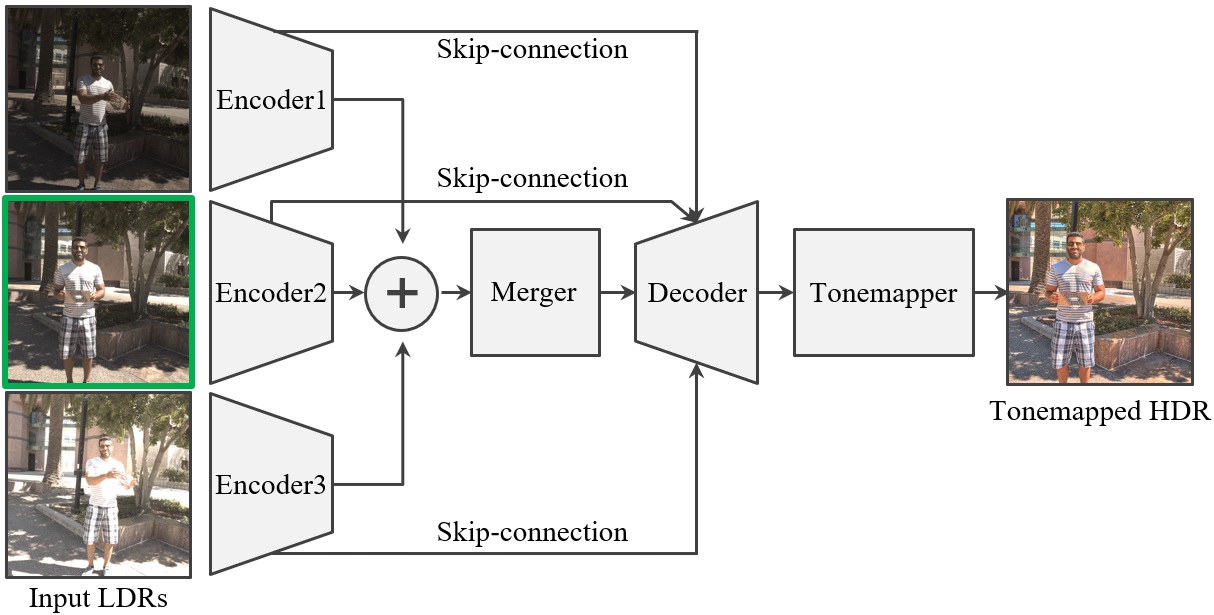} &
				\includegraphics[width=0.24\textwidth]{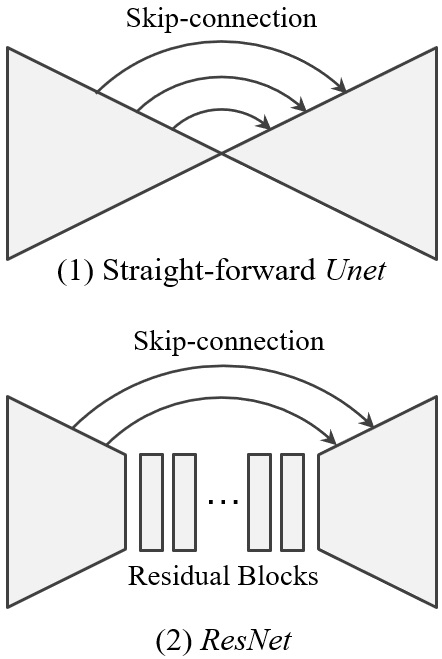} \\
				(a) Network Architecture & (b) Structure
			\end{tabular}
		\end{center}
		\caption{Our framework is composed of three components: encoder, merger and decoder. Different exposure inputs are passed to different encoders, and concatenated before going through the merger and the decoder. We experimented with two structures, \textit{Unet} and \textit{ResNet}. We use skip-connections between the mirrored layers. The output HDR of the decoder is tonemapped before it can be displayed. }
		\label{fig:framework}
	\end{figure}
	
	\section{Approach}
	We formulate the problem of HDR imaging as an image translation problem. Similar to~\cite{LearningHDR}, given a set of LDR images $\{I_1, I_2, ..., I_k\}$, we define a reference image $I_r$. In our experiments, we use three LDRs, and set the middle exposure shot as reference. The same network can be extended to deal with more LDR inputs, and possibly with any specified reference image. We provide results in Section~\ref{sec:diff_ref} to substantiate such robustness. 
	
	Specifically, our goal is to learn a mapping from a stack of LDR images $\{I_1, I_2, I_3\}$ to a ghost-free HDR image $H$ that is aligned with the reference LDR input $I_r$ (same as $I_2$), and contains the maximum possible HDR contents. These contents are either obtained directly from LDR inputs, or from hallucinations when they are completely missing. We focus on handling large foreground motions, and assume the input LDR images, which are typically taken in a burst, have small background motions. 
	
	\subsection{Network Architecture}
	We capitalize on a translation network to learn such a mapping. As shown in Fig.~\ref{fig:framework}, our framework is essentially a symmetric encoder-decoder architecture, with two variants, \textit{Unet} and \textit{ResNet}. 
	
	\textit{Unet}~\cite{unet} is a common tool for translation learning. It is essentially an encoder-decoder architecture, with skip-connections that forward the output of the encoder layer (conv) directly to the input of the corresponding decoder layer (deconv) through channel-wise concatenation. In recent image translation works, such as~\cite{pix2pix}, \textit{Unet} has been demonstrated to be powerful in a wide range of tasks. However, unlike~\cite{pix2pix} where \textit{Unet} was used in an adversarial setting, we may not need a discriminator network in HDR imaging, because the mapping from LDR to HDR is relatively easy to learn, compared to other scenarios in~\cite{pix2pix}, where the two images domains are much more distinct, such as \textit{edge} $\leftrightarrow$ \textit{photo}.
	
	In addition to simple \textit{Unet}, we also experimented with another structure, \textit{ResNet}, similar to \textit{Image Transformation Networks} proposed in~\cite{perceptual_loss}, which simply replaces the middle layers with residual blocks~\cite{ResNet}. Similar structure is also used in recent translation works~\cite{CycleGAN}. In this paper, we name the this structure \textit{ResNet}, as opposed to the previous one, \textit{Unet}. We compare their performance in later sections. 
	
	The overall architecture can be conceptually divided into three components: encoder, merger and decoder. Since we have multiple exposure shots, intuitively we may have separate branches to extract different types of information from different exposure inputs. Instead of duplicating the whole network, which may defer the merging, we separate the first two layers as encoders for each exposure inputs. After extracting the features, the network learns to merge them, mostly in the middle layers, and to decode them into an HDR output, mostly in the last few layers. 
	
	\subsection{Processing Pipeline and Loss Function}
	Given a stack of LDR images, if they are not in RAW format, we first linearize the images using the estimated inverse of Camera Response Function (CRF)~\cite{Grossberg03}, which is often referred to as radiometric calibration. We then apply gamma correction to produce the input to our system. 
	
	Although this process is technically important in order to recover the accurate radiance map, in practice, our system could also produce visually plausible approximation without radiometric calibration, such as examples shown in Fig.~\ref{fig:phone}. This is because the gamma function can be a rough approximation of the CRF. 
	
	We denote the set of input LDRs by $\mathcal{I} = \{I_1, I_2, I_3\}$, sorted by their exposure biases. We first map them to $\mathcal{H} = \{H_1, H_2, H_3\}$ in the HDR domain. We use simple gamma encoding for this mapping: 
	\begin{equation}
	\label{eq:gamma_encoding}
	\begin{split}
	H_i = \frac{I^\gamma_i}{t_i}, \gamma > 1
	\end{split}
	\end{equation}
	where $t_i$ is the exposure time of image $I_i$. Note that we use $H$ to denote the target HDR image, and $H_i$ to denote the LDR inputs mapped to HDR domain. The values of $I_i$, $H_i$ and $H$ are bounded between $0$ and $1$. 
	
	We then concatenate $\mathcal{I}$ and $\mathcal{H}$ channel-wise into a 6-channel input and feed it directly to the network. This is also suggested in~\cite{LearningHDR}. The LDRs facilitate the detection of misalignments and saturation, while the exposure-adjusted HDRs improve the robustness of the network across LDRs with various exposure levels. Our network $f$ is thus defined as: 
	\begin{equation}
	\label{eq:network}
	\hat{H} = f(\mathcal{I}, \mathcal{H})
	\end{equation}
	where $\hat{H}$ is the estimated HDR image, and is also bounded between $0$ and $1$. 
	
	Since HDR images are usually displayed after tonemapping, we compute the loss function on the tonemapped HDR images, which is more effective than directly computed in the HDR domain. In~\cite{LearningHDR} the author proposed to use $\mu$-law, which is commonly used for range compression in audio processing: 
	\begin{equation}
	\label{eq:tonemapper}
	\mathcal{T}(H) = \frac{\log(1+\mu H)}{\log(1+\mu)}
	\end{equation}
	where $H$ is an HDR image, and $\mu$ is a parameter controlling the level of compression. We set $\mu$ to $5000$. Although there are other powerful tonemappers, most of them are typically complicated and not fully differentiable, which makes them not suitable for training a neural network. 
	
	Finally, our loss function is defined as: 
	\begin{equation}
	\label{eq:loss}
	\mathcal{L}_\mathit{Unet} = \|\mathcal{T}(\hat{H}) - \mathcal{T}(H)\|_2
	\end{equation}
	where $H$ is the ground truth HDR image. 
	
	\section{Datasets}
	We used the dataset provided by~\cite{LearningHDR} for training and testing. Although other HDR datasets are available, many of them either do not have ground truth HDR images, or contain only a very limited number of scenes. This dataset contains 89 scenes with ground truth HDR images. As described in~\cite{LearningHDR}, for each scene, 3 different exposure shots were taken while object was moving, and another 3 shots were taken while object remained static. The static sets are used to produce ground truth HDR with reference to the medium exposure shot. This medium exposure reference shot then replaces the medium exposure shot in the dynamic sets. All images are resized to $1000\times1500$. Each set consists of LDR images with exposure biases of $\{-2.0, 0.0, +2.0\}$ or $\{-3.0, 0.0, +3.0\}$. We also tested our trained models on Sen's dataset~\cite{Sen12} and Tursun's dataset~\cite{Tursun15,Tursun16}.

	\subsection{Data Preparation}
	To focus on handling foreground motions, we first align the background using simple homography transformation, which does not introduce artifacts and distortions. This makes the learning more effective than directly trained without background alignment. Comparison and discussion are provided in Section~\ref{sec:bg_align}. 
	
	\subsection{Data Augmentation and Patch Generation}
	The dataset was split into 74 training examples and 15 testing examples by~\cite{LearningHDR}. For the purpose of efficient training, instead of feeding the original full-size image into our model, we crop the images into $256 \times 256$ patches with a stride of 64, which produces around 19000 patches. We then perform data augmentation (flipping and rotation), further increasing the training data by 8 times. 
	
	In fact, a large portion of these patches contain only background regions, and exhibit little foreground motions. To keep the training focused on foreground motions, we detect large motion patches by thresholding the structural similarity between different exposure shots, and replicate these patches in the training set. 
	
	\begin{table}[t]
		\caption{Comparison of average running time on the test set under CPU environment. }
		\centering
		\begin{tabular}{c|c|c|c|c|c|c}
			\hline
			& Sen~\cite{Sen12} & Hu~\cite{Hu13} & Kalantari~\cite{LearningHDR} & HDRCNN~\cite{hdrcnn} & Ours \textit{Unet} & Ours \textit{ResNet} \\
			\hline
			Time (s) & 261 & 137 & 72.1 & 12.6 & 11.9 & 14.7\\
			\hline
		\end{tabular}
		\label{tab:time}
	\end{table}

	\begin{table}[t]
		\caption{Quantitative comparisons of the results on Kalantari's test set~\cite{LearningHDR}. The first two rows are PSNR/SSIM computed using tonemapped outputs and ground truth, and the following two rows are PSNR/SSIM computed using linear images and ground truth. The last row is HDR-VDP-2~\cite{HDR-VDP-2} sores. All values are the average across 15 testing images in the original test set. }
		\centering
		\begin{tabular}{c|c|c|c|c|c}
			\hline
			& Sen~\cite{Sen12} & Hu \cite{Hu13} & Kalantari~\cite{LearningHDR} & Ours \textit{Unet} & Ours~\textit{ResNet} \\
			\hline
			PSNR-T & 40.80 & 35.79 & \textbf{42.70} & 40.81 & 41.65\\
			\hline
			SSIM-T & 0.9808 & 0.9717 & \textbf{0.9877} & 0.9844 & 0.9860\\
			\hline
			PSNR-L & 38.11 & 30.76 & \textbf{41.22} & 40.52 & 40.88\\
			\hline
			SSIM-L & 0.9721 & 0.9503 & 0.9845 & 0.9837 & \textbf{0.9858}\\
			\hline
			HDR-VDP-2 & 59.38 & 57.05 & 63.98 & 64.88 & \textbf{64.90}\\
			\hline
		\end{tabular}
		\label{tab:compare_quan}
	\end{table}
	
	\begin{figure}[t]
		\begin{center}
			\includegraphics[width=0.89\textwidth]{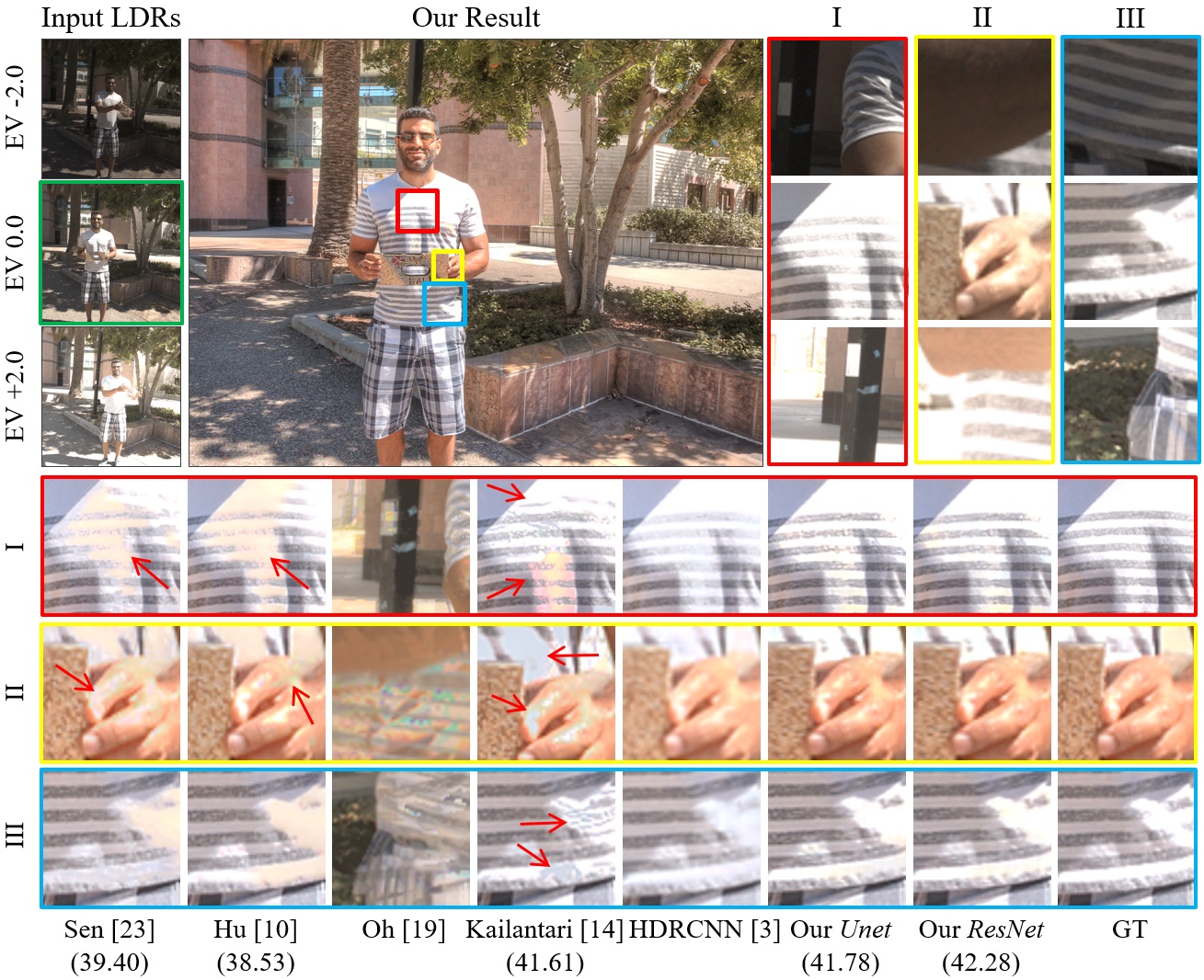}
		\end{center}
		\caption{Comparison against several state-of-the-art methods. In the upper half of the figure, the left column shows in the input LDRs, the middle is our tonemapped HDR result, and the last three columns show three zoomed-in LDR regions marked in the HDR image. The lower half compares the zoomed-in HDR regions of our results against others. The numbers in brackets at the bottom indicate the PSNR of the tonemapped images. Images are obtained from the Kalantari's test set~\cite{LearningHDR}. }
		\label{fig:qualitative}
	\end{figure}
	
	\section{Experiments and Results}
	
	\subsection{Implementation Details}
	We first perform radiometric calibration and map the input LDRs to HDR domain. Each of the resulted radiance maps is channel-wise concatenated with the LDR image respectively, and then separately fed into different encoders. After 2 layers, all feature maps are then concatenated channel-wise for merging. 
	
	The encoding layers are convolution layers with a stride of 2, while the decoding layers are deconvolution layers kernels with a stride of 1/2. The output of the last deconvolution layer is connected to a flat-convolution layer to produce the final HDR. All layers use $5\times5$ kernels, and are followed by batch normalization (except the first layer and the output layer) and leaky ReLU (encoding layers) or ReLU (decoding layers). The channel numbers are doubled each layer from 64 to 512 during encoding and halved from 512 to 64 during decoding. 
	
	For \textit{Unet} structure, $256\times256$ input patches are passed through 8 encoding layers to produce a $1\times1\times512$ block, followed by 8 decoding layers plus an output layer to produce a $256\times256$ HDR patch. Our \textit{ResNet} is different only in that after 3 encoding layers, the $32\times32\times256$ block is passed through 9 residual blocks with $3\times3$ kernels, followed by 3 decoding layers and an output layer. 
	
	\begin{figure}[t]
		\begin{center}
			\includegraphics[width=0.99\textwidth]{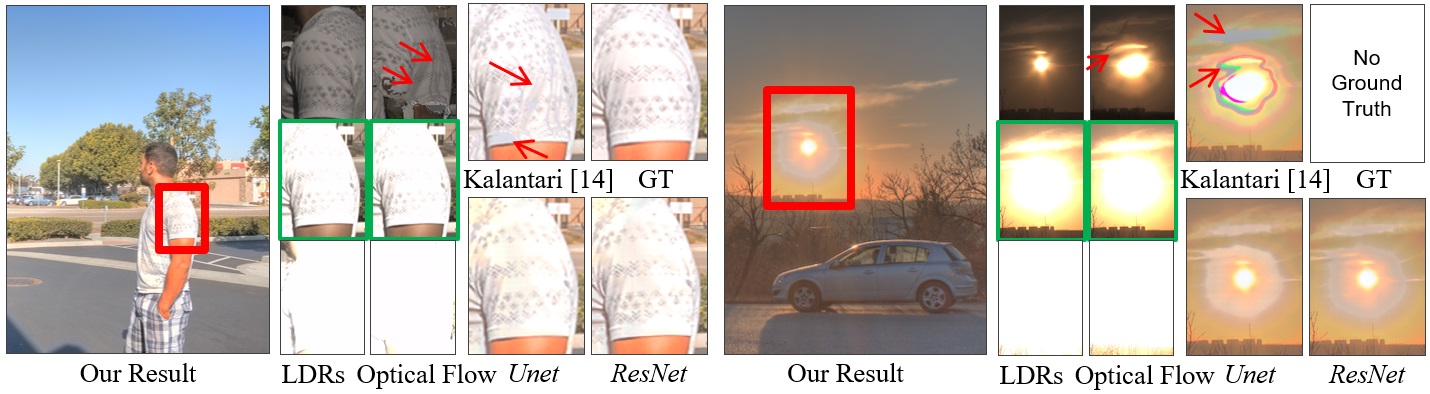}
		\end{center}
		\caption{Comparison against flow-based method~\cite{LearningHDR}. Images are obtained from the Kalantari's dataset~\cite{LearningHDR} and Tursun's dataset~\cite{Tursun15,Tursun16}. }
		\label{fig:optical_flow}
	\end{figure}
	
	\begin{figure}[t]
		\begin{center}
			\includegraphics[width=\textwidth]{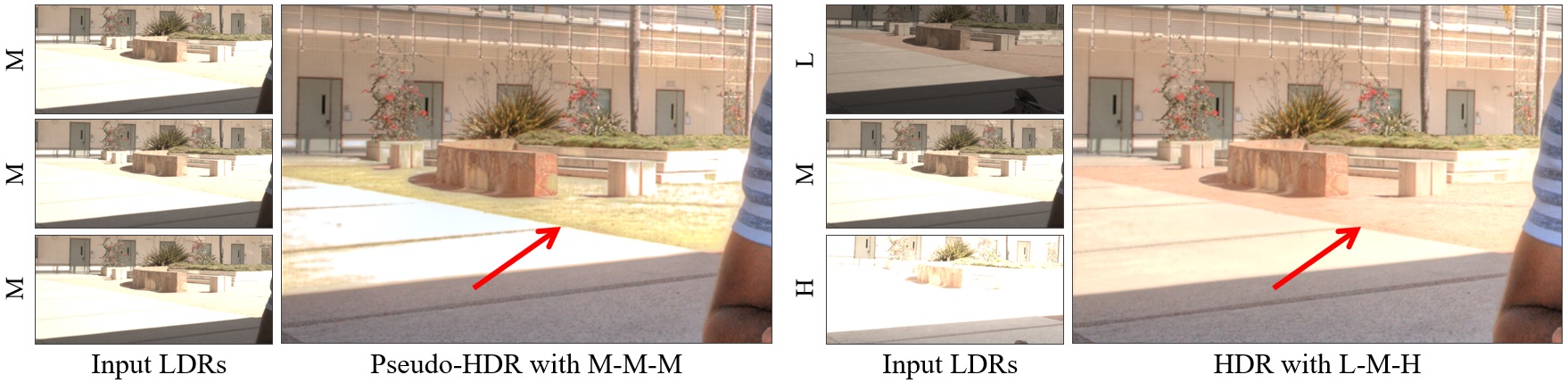}
		\end{center}
		\caption{Example of hallucination. The left is generated using only medium exposure shot, and the right is generated using low, medium and high exposure shots. Images are obtained from the Kalantari's dataset~\cite{LearningHDR}. }
		\label{fig:hallucination1}
	\end{figure}
	
	\begin{figure}[ht]
		\begin{center}
			\includegraphics[width=0.75\textwidth]{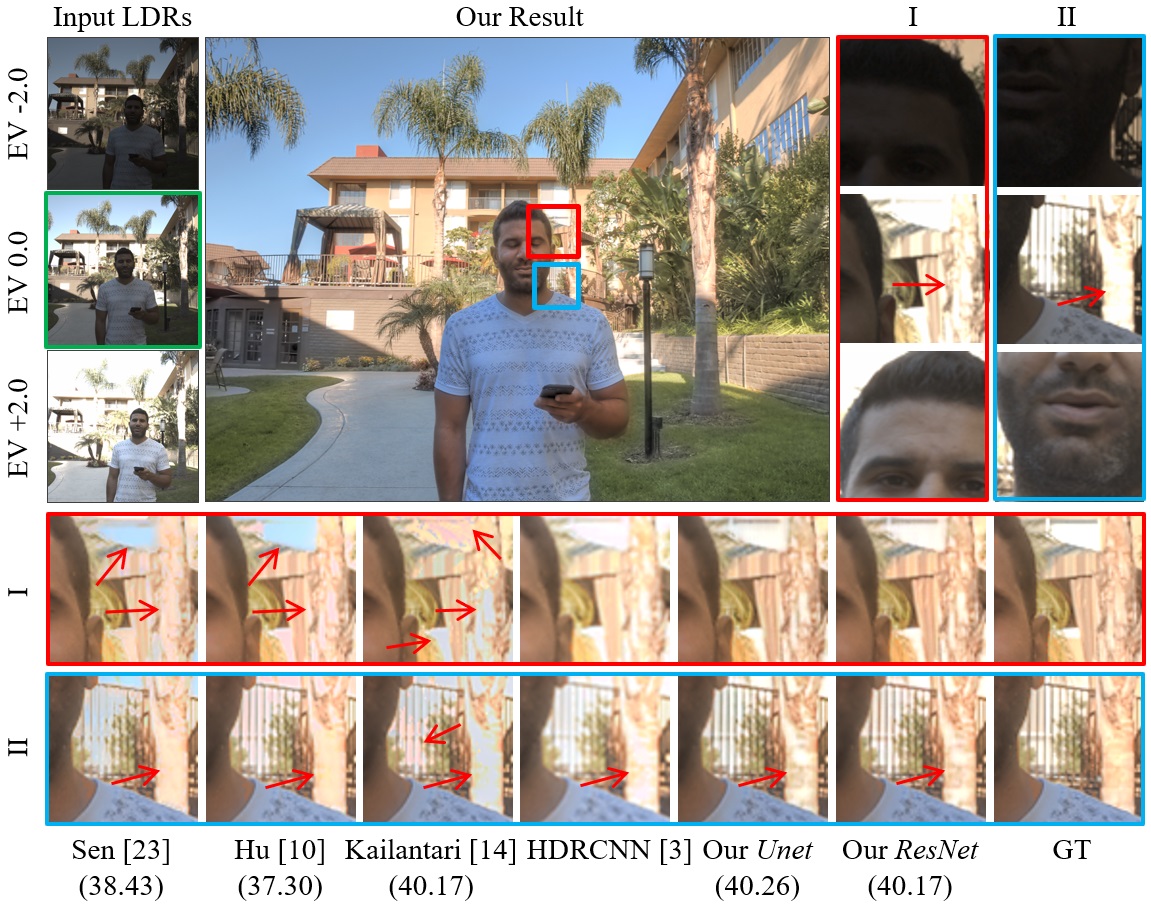}
		\end{center}
		\caption{Comparison of hallucinated details. Our network hallucinates the missing trunk texture, while others may fail. Images are obtained from the Kalantari's dataset~\cite{LearningHDR}. }
		\label{fig:hallucination2}
	\end{figure}
	
	\subsection{Running Time}
	We report running time comparison with other methods in Table~\ref{tab:time}. Although our network is trained with GPU, other conventional optimization methods are optimized with CPU. For fair comparison, we evaluated all methods under CPU environment, on a PC with i7-4790K (4.0GHz) and 32GB RAM. We tested all methods using 3 LDR images of size $896\times1408$ as input. Note that the optical flow alignment used in~\cite{LearningHDR} takes 59.4s on average. When run with GPU (Titan X Pascal), our \textit{Unet} and \textit{ResNet} take \textcolor{red}{0.225s} and \textcolor{red}{0.239s} respectively. 
	
	\subsection{Evaluation and Comparison}
	\label{sec:eval_compare}
	We perform quantitative and qualitative evaluations, and compare results with the state-of-the-art methods, including two patch-based methods~\cite{Sen12,Hu13}, motion rejection method~\cite{Oh15}, the flow-based method with CNN merger~\cite{LearningHDR}, and the single image HDR imaging~\cite{hdrcnn}. For all methods, we used the codes provided by the authors. Note that all the HDR images are displayed after tonemapping using \textit{Photomatix}~\cite{photomatix}, which is different from the tonemapper used in training. 
	
	\subsubsection{Quantitative Comparison}
	\label{sec:quan_compare}
	We compute the PSNR and SSIM scores between the generated HDR and the ground truth HDR, both before and after tonemapping using $\mu$-law. We also compute the HDR-VDP-2~\cite{HDR-VDP-2}, a metric specifically designed for measuring the visual quality of HDR images. For the two parameters used to compute the HDR-VDP-2 scores, we set the diagonal display size to 24 inches, and the viewing distance to 0.5 meter. We did not compare with~\cite{Oh15} and~\cite{hdrcnn} quantitatively, since the former is optimized for more than 5 LDR inputs and the latter produces unbounded HDR results. 
	
	Table~\ref{tab:compare_quan} shows quantitative comparison of our networks against the state-of-the-art methods. Note that all results are calculated on the Kalantari's test set~\cite{LearningHDR}. While~\cite{LearningHDR} results in slightly higher PSNR scores, our methods result in comparable SSIM scores and slightly higher HDR-VDP-2 scores. Besides, \textit{ResNet} seems to yield higher scores than \textit{Unet}. 
	
	\begin{figure}[t]
		\begin{center}
			\includegraphics[width=0.79\textwidth]{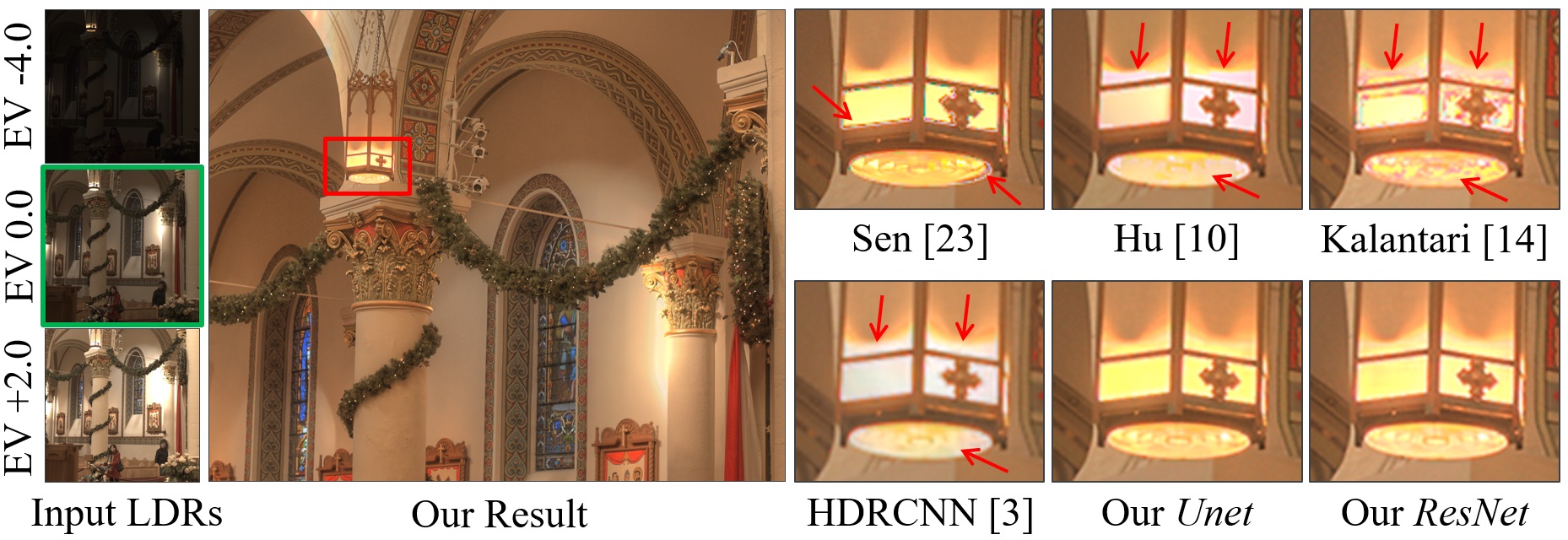}
		\end{center}
		\caption{Comparison of highlight regions. Examples come from the Sen's dataset~\cite{Sen12}. }
		\label{fig:highlight}
	\end{figure}
	
	\begin{figure}[t]
		\begin{center}
			\includegraphics[width=0.7\textwidth]{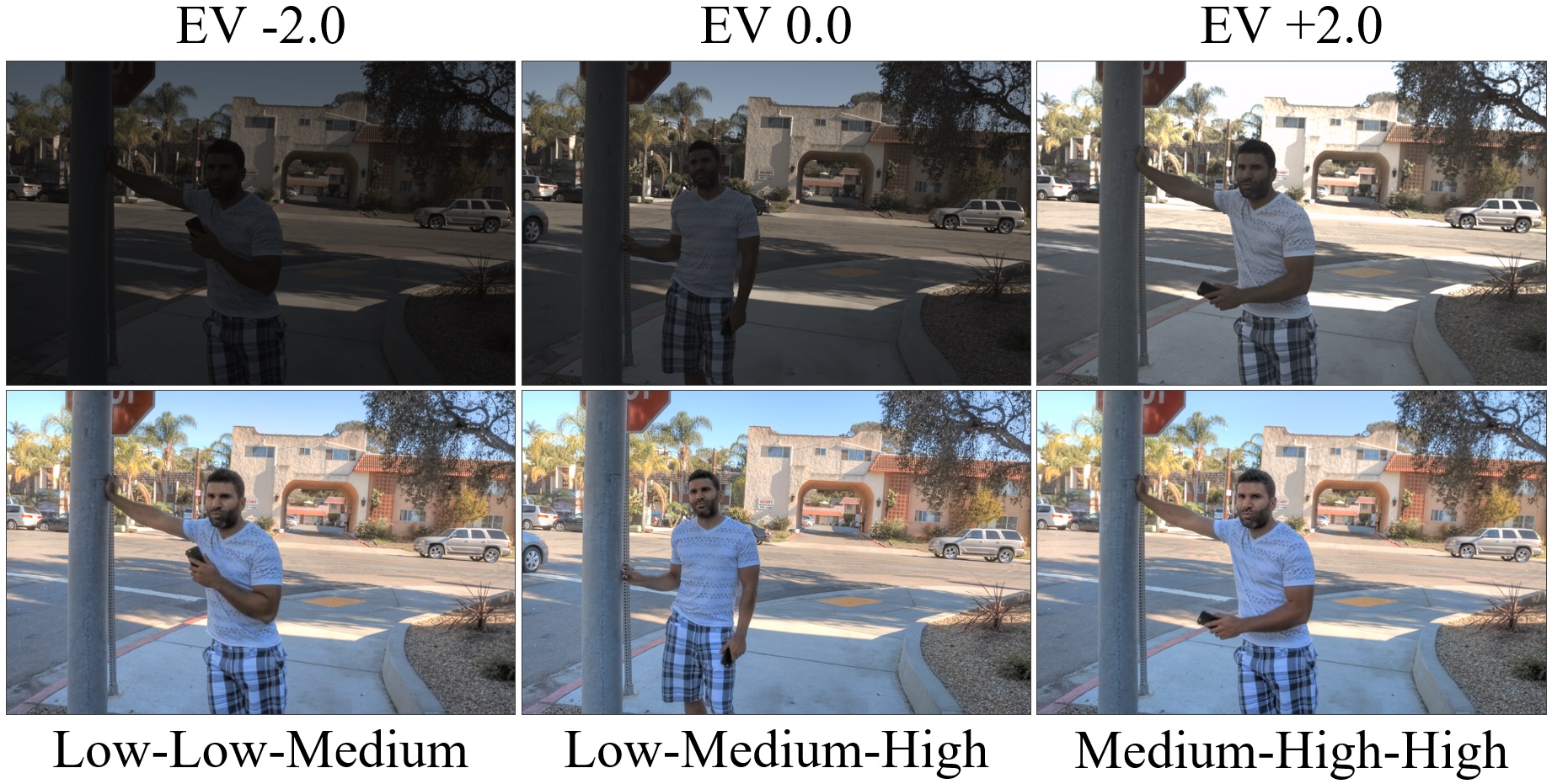}
		\end{center}
		\caption{Results with different reference images. The first row shows three LDR inputs, and the second row shows the corresponding HDR results with reference to each input. }
		\label{fig:diff_ref}
	\end{figure}
	
	\begin{figure}[t]
		\begin{center}
			\includegraphics[width=0.99\textwidth]{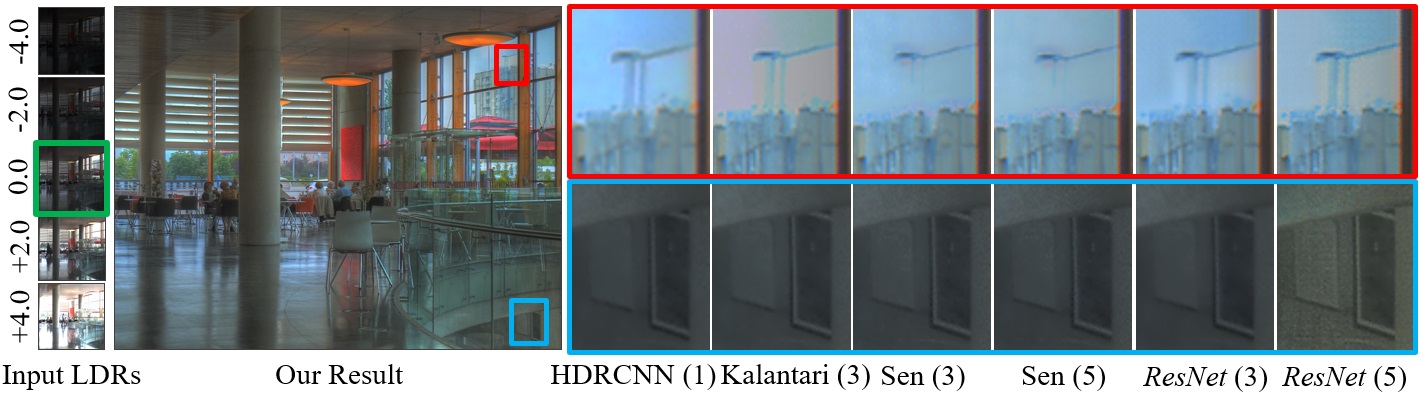}
		\end{center}
		\caption{Results with more input LDRs. The integers in the parentheses indicate the number of LDR images used to generate produce the HDR. }
		\label{fig:5_input}
	\end{figure}
	
	\begin{figure}[t]
		\begin{center}
			\begin{tabular}{c c c}
				\includegraphics[width=0.347\textwidth]{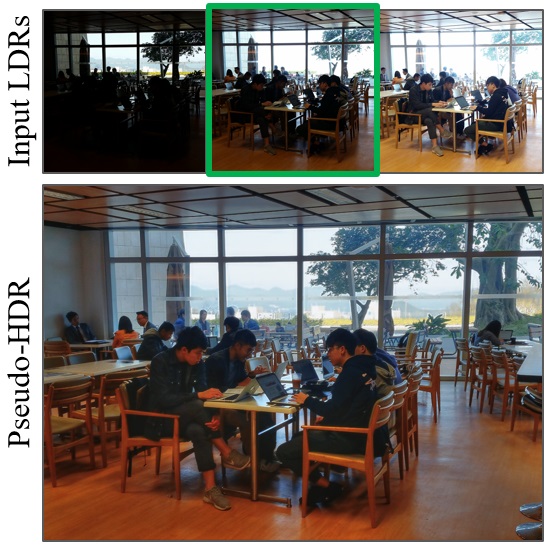} & \includegraphics[width=0.324\textwidth]{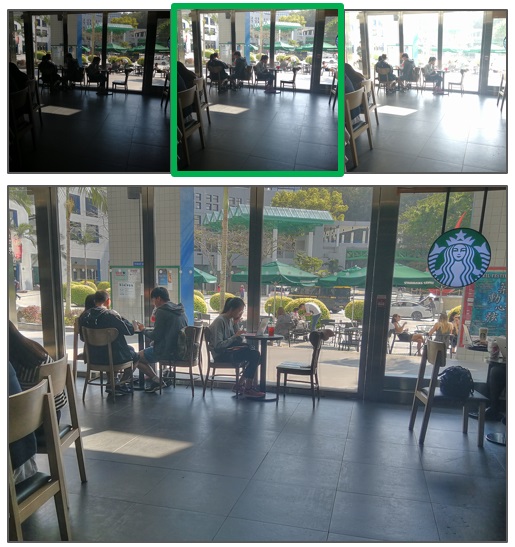} &
				\includegraphics[width=0.324\textwidth]{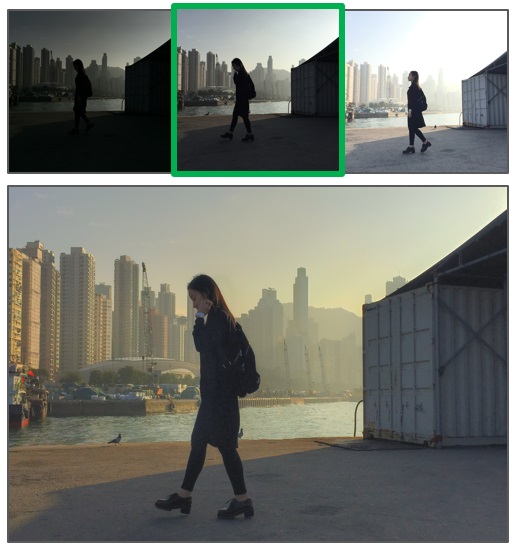} \\
				(a) Samsung Galaxy S5 & (b) Huawei Mate 9 & (c) iPhone 6s
			\end{tabular}
		\end{center}
		\caption{HDR results without radiometric calibration. All examples are novel images taken using cellphones with different CRFs. }
		\label{fig:phone}
	\end{figure}
	
	\subsubsection{Qualitative Comparison}
	Fig.~\ref{fig:qualitative} compares the testing results against state-of-the-art methods. In regions with no object motions, all methods produce decent results. However, when large object motion is present in saturated regions,~\cite{Sen12,Hu13,LearningHDR} tend to produce unsightly artifacts. Flow-based method~\cite{LearningHDR} also produces geometric distortions. Because Oh's method~\cite{Oh15} uses rank minimization, which generally requires more inputs, it results in ghosting artifacts when applied with 3 inputs. Since HDRCNN~\cite{hdrcnn} estimates the HDR image using only one single reference LDR image, it does not suffer from object motions, but tends to produce less sharp results and fail in large saturated regions, as shown in Fig.~\ref{fig:teaser}. Our two networks produce comparably good results, free of obvious artifacts and distortions. In general, \textit{ResNet} seems to consistently outperform \textit{Unet}. 
	
	\subsubsection{Comparison against Flow-Based Method}
	In addition to Fig.~\ref{fig:teaser} and Fig.~\ref{fig:qualitative}, Fig.~\ref{fig:optical_flow} illustrates our advantages over Kalantari's method~\cite{LearningHDR}, where optical flow alignment introduces severe distortions and color artifacts. Our method does not rely on erroneous optical flow, which intrinsically avoids such distortions, and is also much more efficient computationally. 
	
	\subsubsection{Hallucination}
	\label{sec:hallucination}
	One important feature of our method is the capability of hallucinating missing details that are nearly impossible to recover using conventional optimization approaches. As shown in Fig.~\ref{fig:hallucination1}, when given only the medium exposure, our network is able to properly hallucinate the grass texture in the saturated regions. When given also two other exposure shots, our network is able to incorporate the additional information such as the ground texture. 
	
	In Fig.~\ref{fig:hallucination2}, we examine the effectiveness of hallucination, by comparing our results to others with no hallucination. Hallucination can be very useful in dynamic scenes, since contents in over-exposed or under-exposed regions are often missing in all LDRs due to total occlusions caused by object motions. 
	
	\subsubsection{Highlight}
	\label{sec:highlight}
	In addition to Fig.~\ref{fig:optical_flow}, where we show that our method outperforms~\cite{LearningHDR} in highlight regions, Fig.~\ref{fig:highlight} compares our highlight details against others. While other methods often fail to recover details in highlight regions and introduce artifacts and distortions, our method generally works well. Specifically, Hu's method~\cite{Hu13} performs poorly in general at highlight regions, and other methods can only partially recover the details. Kalantari's method~\cite{LearningHDR} tends to introduce evident distortions and color artifacts as shown in Fig.~\ref{fig:highlight}. 
	
	\subsubsection{Different Reference Image}
	\label{sec:diff_ref}
	Fig.~\ref{fig:diff_ref} illustrates another advantage of our image translation formulation: the flexibility in choosing different reference images. Currently this is achieved by re-arranging the input LDRs. For example, using only low and high exposure shots and feeding them to the network in the order of \{Low-Low-Medium\} will result in a pseudo-HDR image with reference to the low exposure shot. Technically, this output does not represent the accurate radiance values, but is perceptually compelling and similar to real HDR images. Our framework may be extended to directly output multiple HDR images with different reference images, if trained in such a fashion, although we do not have appropriate datasets to corroborate this. 
	
	\begin{figure}[t]
		\begin{center}
			\includegraphics[width=0.8\textwidth]{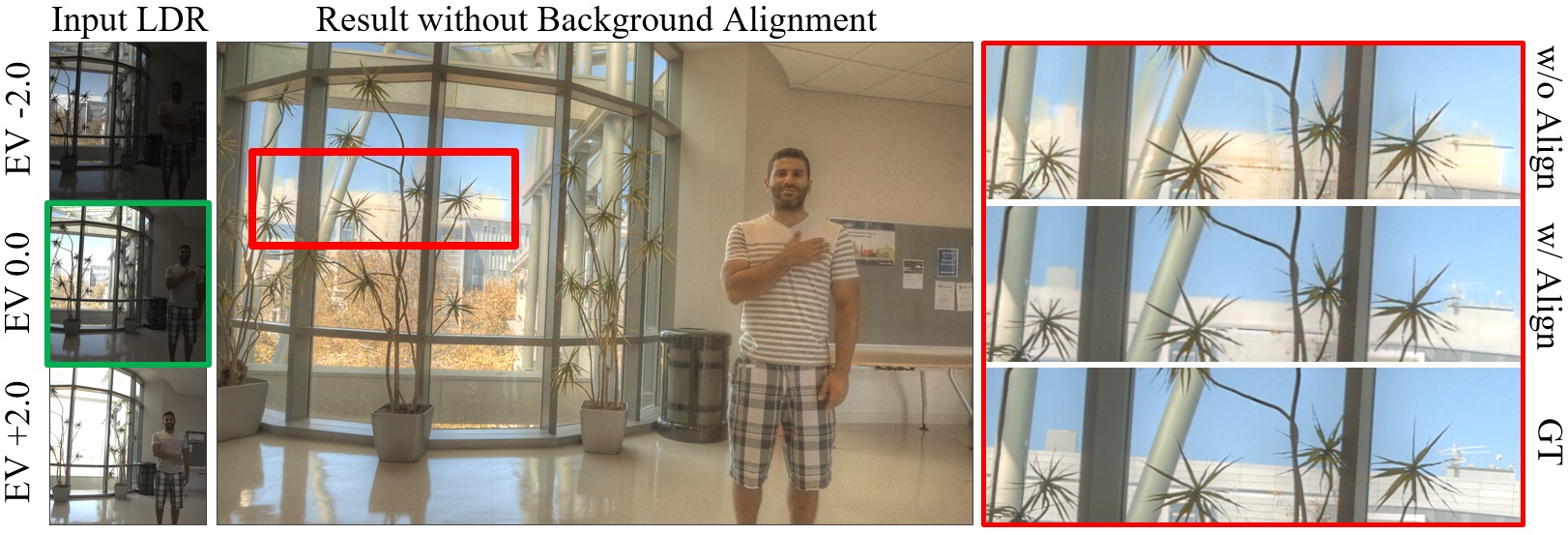}
		\end{center}
		\caption{This example illustrates the effect of background alignment. }
		\label{fig:bg_align}
	\end{figure}
	
	\begin{figure}[t]
		\begin{center}
			\includegraphics[width=\textwidth]{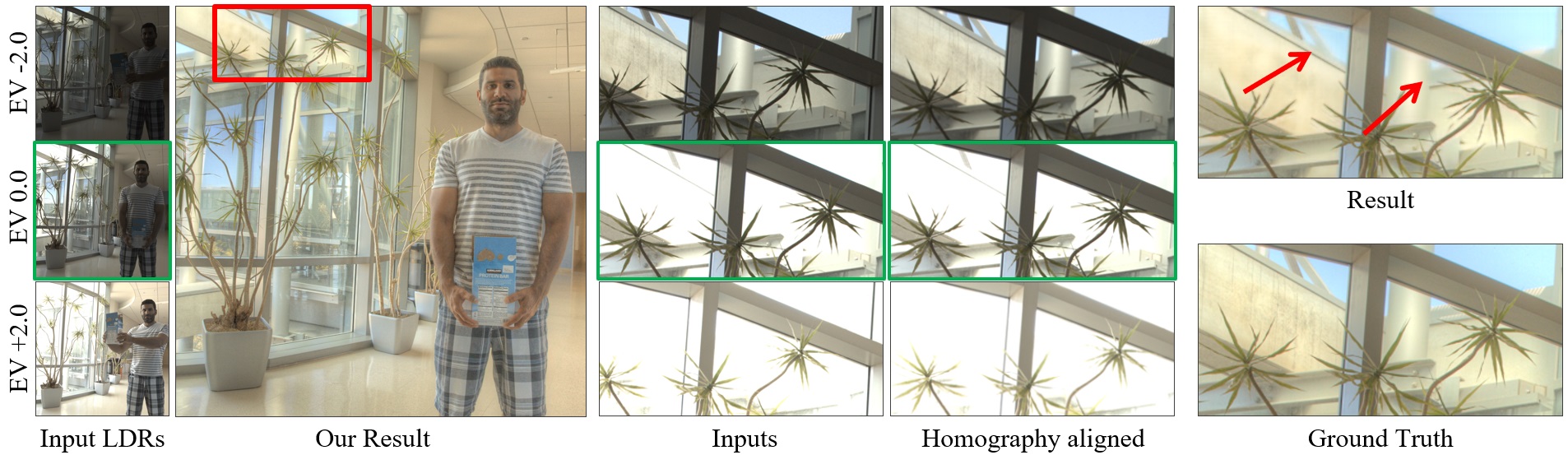}
		\end{center}
		\caption{Blurry results caused by parallax effects, which cannot be resolved by homography transformation. }
		\label{fig:homog_fail}
	\end{figure}
	
	\subsubsection{More Input LDRs}
	\label{sec:5_input}
	Our framework can potentially be extended for supporting more than 3 input LDRs. This is useful, because more LDRs capture more contents and improve the robustness. Although we do not have a suitable dataset to fully explore this, we decided to conduct a brief experiment using Sen's dataset~\cite{Sen12}. We used their produced HDR images as ground truth for training, which are yet to be perfect to be used as ground truth, but sufficient for our purpose of testing such extensibility. Using this dataset, we tested our framework using 5 LDR inputs. Fig.~\ref{fig:5_input} compares our results with others. Interestingly, while Sen's~\cite{Sen12} results using 5 inputs do not seem to be clearly better than those using 3 inputs, in our results, the details in saturated and under-exposed regions are markedly improved by using more input LDRs. 
	
	\subsubsection{Cellphone Example}
	We also tested our model on novel cellphone images for proof of practicality, shown in Fig.~\ref{fig:phone}. Our network produces good results in various kinds of settings. The input images were captured using different cellphones with different camera response functions. It is worth noting that when producing these pseudo-HDR examples, we did not perform radiometric calibration. This again demonstrates the robustness of our network. 
	
	\subsection{Discussion on Background Alignment}
	\label{sec:bg_align}
	In all our experiments and comparisons, since we are focused on handling large foreground motions, we align the backgrounds of the LDR inputs using homography transformation. Without background alignment, we found that our network tends to produce blurry edges where background is largely misaligned, as shown in Fig.~\ref{fig:bg_align}. This can be due to the confusion caused by the background motion, which CNN is generally weak at dealing with. However, such issues can be easily resolved using simple homography transformation that almost perfectly aligns the background in most cases. Recall that in practice, the LDR inputs can be captured in a burst within a split second using nowadays handheld devices. 
	
	Nevertheless, homography is not always perfect. One particular case where homography may not produce perfect alignment is the existence of parallax effects in saturated regions. The final HDR output may be blurry. See Fig.~\ref{fig:homog_fail}. 
	
	\section{Conclusion and Future Work}
	In this paper, we demonstrate that the problem of HDR imaging can be formulated as an image translation problem and tackled using deep CNNs. We conducted extensive quantitative and qualitative experiments to show that our non-flow-based CNN approach outperforms the state-of-the-arts, especially in the presence of large foreground motions. In particular, our simple translation network intrinsically avoids distortions and artifacts produced by erroneous optical flow alignment, and is computationally much more efficient. Furthermore, our network can hallucinate plausible details in largely saturated regions with large foreground motions, and recovers highlight regions better than other methods. Our system can also be easily extended with more inputs, and with different reference images, not limited to the medium exposure LDR. It is also robust across different inputs, including images that are not radiometrically calibrated. 
	
	While our advantages are clear, it is yet to be a perfect solution. We also observe challenges of recovering massive saturated regions with minimal number of input LDRs. In the future, we would attempt to incorporate high-level knowledge to facilitate such recovery, and devise a more powerful solution. 
	
	\subsubsection*{Acknowledgement}This work was supported in part by Tencent Youtu. 
	
	%
	%
	%
	\bibliographystyle{splncs04}
	\bibliography{hdr}
\end{document}